\pdfoutput=1
\documentclass[11pt]{article}

\usepackage[]{acl}

\usepackage{times}
\usepackage{microtype}
\usepackage{subfigure}
\usepackage{graphics}

\usepackage{booktabs} % For formal tables
\usepackage{latexsym}
\usepackage{amsmath}
\usepackage{xspace}
\usepackage{amssymb}
\usepackage{graphicx}
\usepackage{multirow}
\usepackage{enumitem}
\usepackage{ulem}
\usepackage{float}
\usepackage{makecell}

\usepackage{appendix}

\usepackage[T1]{fontenc}
\usepackage[utf8]{inputenc}

\usepackage{microtype}

\newcommand{\ie}{\textit{i.e.,}\xspace}
\newcommand{\eg}{\textit{e.g.,}\xspace}

\newcommand{\paratitle}[1]{\vspace{0.8ex}\noindent \textbf{#1}}
\newcommand{\camera}{CameraReview\xspace}
\newcommand{\compsentliu}{CompSent-08\xspace}

\newcommand{\modelname}{GCRE-GPT\xspace}

\title{GCRE-GPT: A Generative Model for Comparative Relation Extraction}
\author{Yequan Wang$^{1}$, Hengran Zhang$^{2,3}$, Aixin Sun$^{4}$, Xuying Meng$^{2}$\\
$^{1}$Beijing Academy of Artificial Intelligence, Beijing, China\\
$^{2}$Institute of Computing Technology, Chinese Academy of Sciences, Beijing, China\\
$^{3}$University of Chinese Academy of Sciences, Beijing, China\\
$^{4}$School of Computer Science and Engineering, Nanyang Technological University, Singapore\\
tshwangyequan@gmail.com, axsun@ntu.edu.sg,  \{zhanghengran22z,mengxuying\}@ict.ac.cn
}

\usepackage{threeparttable}
\begin{document}
\maketitle

\begin{abstract}
Given comparative text, comparative relation extraction aims to extract two targets (\eg two cameras) in comparison and the aspect they are compared for (\eg image quality).  The extracted comparative relations form the basis of further opinion analysis.
%of finding the relative opinion preference.
% It is important for the comparative sentiment classification. 
Existing solutions formulate this task as a sequence labeling task, to extract targets and aspects. However, they cannot directly extract comparative relation(s) from text. In this paper, we show that comparative relations can be directly extracted with high accuracy, by generative model. Based on GPT-2, we propose a Generation-based Comparative Relation Extractor (GCRE-GPT). Experiment results show that \modelname achieves state-of-the-art accuracy on two datasets. %, and our model is able to extract multiple relations in a sentence.
% Comparative relations extraction can simplify comparative sentence and it's crucial for opinion mining. 
% Current solutions mainly use pipelined models. But pipelined models have error propagation problem. 
\end{abstract}

%===========================
\section{Introduction}
%===========================

Comparison is a common linguistic expression. \citet{DBLP:conf/emnlp/KesslerK13} reports that $10\%$ of texts contain at least one comparison. A typical comparative relation contains two targets and one aspect, \eg ``D80 vs. D70 in terms of weight'' where the two targets are two cameras and the aspect to compare is weight.   
An extracted comparative relation is the basis for opinion mining in further analysis.
Hence, it is crucial to accurately extract comparative relations, to support product/service comparison to better inform consumers and enterprises.   

%Comparative texts on various products, enterprises and consumers can make more competitive and scientific decisions.
% Comparative relation extraction aims to extract all comparative relations.

% (\ie entities and corresponding aspects) and analyze comparative sentence opinion towards them. 

% \input{tables/task_representation}
% In existing studies, comparative relation has various expressions, which are shown in Table~\ref{tab:task_representation}. Most of the works considers opinion words as part of the comparison relationship. But opinion words isn't necessary for comparative opinion mining ~\cite{DBLP:conf/argmining/PanchenkoBFHB19}. In addition, extracting more elements, the task will be more complex.

%To simply express the question, here we make a unified definition for comparative relation extraction. 

Given a piece of text (\eg a sentence), denoted by $S = [w_1, w_2, \dots, w_n]$, the task  of \textit{comparative relation extraction} is to predict the comparative relation(s) expressed in the given text. Each relation is a 3-tuple: $(t_1, t_2, a)$, where $t_1$ and $t_2$ are targets in comparison, and $a$ denotes the corresponding aspect for which the two targets are compared. In some studies, the two targets are further classified into subject and object~\cite{DBLP:conf/emnlp/LiuXY21,DBLP:conf/cikm/AroraAGP17,DBLP:conf/emnlp/KesslerK13,DBLP:conf/aaai/JindalL06,DBLP:journals/comsis/WangXWHL17}. In our problem definition, we do not distinguish such roles between the two targets.

% Consider the task is to be conducted on camera reviews. Given a sentence ``\textit{I think the sound quality of this phone is slightly better than my samsung phone , the motorolla t720 series.}'',  we can extract comparative relations, \ie \textit{(this phone, my samsung phone,  sound quality), and (this phone,the motorolla t720 series,  sound quality)}.

Existing solutions for comparative relation extraction use rules or sequence labeling models. 
\citet{DBLP:conf/aaai/JindalL06} extract pre-defined comparative tuples using rules based on opinion words. \citet{DBLP:conf/emnlp/KesslerK13} recognize comparative predicate by semantic role labeling method. 
Furthermore, CRF~\cite{DBLP:journals/comsis/WangXWHL17} and LSTM~\cite{DBLP:conf/cikm/AroraAGP17} have also been used to extract the comparative elements.
\citet{DBLP:conf/emnlp/LiuXY21} use sequence labeling method to obtain elements, then use Cartesian product to compute all relations. Here, multiple relations in a single sentence may be combined by Cartesian product. Nevertheless, multiple relations in long texts (\eg a paragraph) may not be logically combined. Although sequence labeling models perform well at sentence level, they cannot directly extract relation tuples. Instead, they extract targets and aspects and combine them into relation tuples. The permutation and combination of elements bring in a lot of complexity.
More importantly, many reviews in real life contain more than one relation.
This calls for a fundamentally different approach to extract comparative relations.

Pre-trained language models, \eg GPT~\cite{radford2019language} and BERT~\cite{DBLP:conf/naacl/DevlinCLT19}, have enabled generative models to achieve promising results on many tasks. Examples include  machine translation~\cite{DBLP:journals/taslp/ZhangLSZX021,DBLP:conf/acl/TuLLLL16}, question answer~\cite{DBLP:conf/emnlp/DuanTCZ17}, and generation-based dialog system~\cite{DBLP:conf/acl/TuLC0W022}. In this paper, we make the first attempt to extract comparative relations with generative model.
%we take the first step using the generation way to solve the difficulty caused by multiple relations. 
Specifically, we propose \textbf{G}eneration-based \textbf{C}omparative \textbf{R}elation \textbf{E}xtractor based on \textbf{GPT}-2, namely \modelname, to extract comparative relations from text. 
%Aixin: I do not understand the next sentence, "rule" is strange here. 
We represent a comparative relation tuple in a piece of text, and train our model to generate such texts, based on an input sentence. Experiments show that \modelname achieves state-of-the-art performance. Further, \modelname performs best on texts containing multiple relations.

%Aixin: this part is repeating, suggest to remove
%To summarize, we make the first attempt to extract comparative relations by \modelname, which is based on generation method. More importantly, the proposed method overcomes the difficulty caused by multiple relations fundamentally. Experiment results show that our proposed model achieves state-of-the-art performance, especially in texts with multiple relations.

% for comparative relation extraction, using generative tasks to accomplish comparative relation extraction. There are two restrictions during generation: Format Restriction and  Prompt words injection.

%===========================
\section{Related Work}
%===========================
Most existing studies cast the task of comparative relation extraction as a task of  comparative element extraction. Relations are then built on top of the extracted elements.  Accordingly, many solutions are based on sequence labeling models~\cite{DBLP:conf/aaai/JindalL06, DBLP:conf/emnlp/KesslerK13, DBLP:journals/comsis/WangXWHL17, DBLP:conf/cikm/AroraAGP17, DBLP:conf/emnlp/LiuXY21}. These models are fundamentally different from ours, as we adopt the generative approach with pre-trained language models.

As a well known pre-training language model, BERT~\cite{DBLP:conf/naacl/DevlinCLT19} has contributed to significant performance increase on various NLP tasks. Based on BERT, \citet{DBLP:conf/emnlp/LiuXY21} propose a multiple stage model to analyze comparative opinion. The comparative elements are firstly extracted, followed by building comparative relations. We instead adopt a generative approach, built on top of GPT-2. GPT-2~\cite{radford2019language} adopts generic transformer structure~\cite{DBLP:conf/acl/ZhangSGCBGGLD20} and encodes text from left to right. The auto-regressive and unidirectional structure make it a strong performer on generation tasks. 

% \citet{DBLP:journals/corr/abs-2103-10360} reformulate downstream tasks as the blank-filling generation, which surprisingly outperforms other Large-scale language models.
% T5~\cite{} converts all NLP tasks to generation tasks. In detail, T5 uses prefix descriptions to fit various tasks. For example, the translation task adds ``translate English to German'' as prompt words to translate English.  So it also can change the comparative relation extraction task to a sequence-to-sequence task. 

%===========================
\section{\modelname}
\label{sec:model}
\begin{figure}
    \centering
    \includegraphics[scale=0.45]{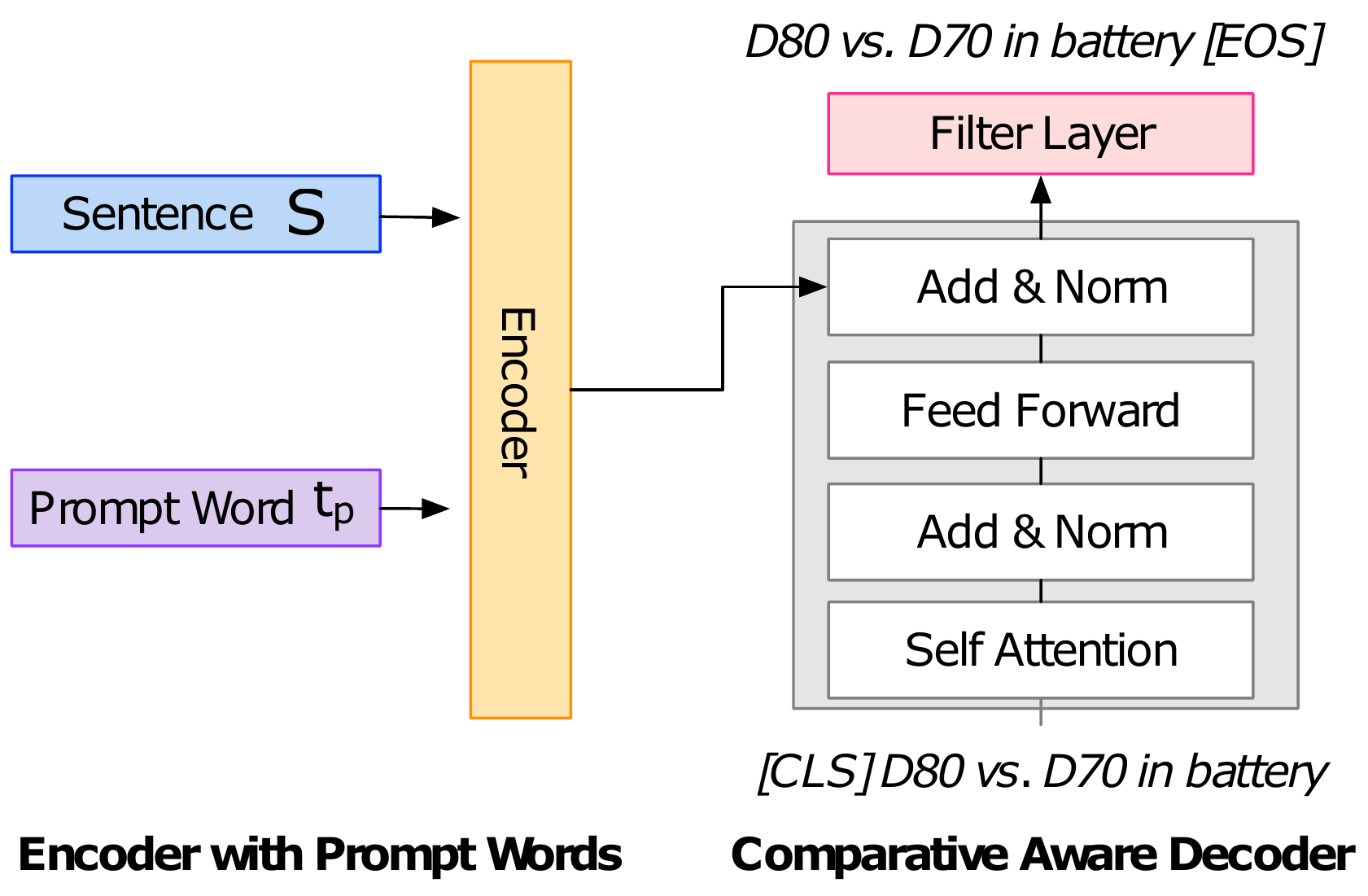}
%     \caption{
%   An instance of prompt words injection.  ``sentences:'' and ``relation:'' are prompt words and prompt words can be changed.  When training the model, the sentence and relations are separated by prompt words. When testing, sentence and prompt words as input and model generate relations. $c$ is the hidden vector. $p^c$ is probability distribution. 
%     }
    \caption{The architecture of \modelname.}
    \label{fig:model}
    % \vspace{-2ex}
\end{figure}

%===========================
The architecture of the proposed \modelname is depicted in Figure~\ref{fig:model}. 
Rooted in GPT-2, \modelname has two main modules: (i) encoder with prompt words, and (ii) comparative aware decoder.
Next, we detail the design of \modelname and its optimization objective.

% The  \textbf{G}enerated-based \textbf{C}omparative \textbf{R}elation \textbf{E}xtractor based on \textbf{GPT}~(\modelname) innovatively use prompt words injection.  Figure \ref{fig:model}  depicts  the  architecture  of \modelname. Based on GPT-2 model, \modelname has three modules:\textit{Content Restriction},  \textit{Format Restriction} and \textit{Prompt Words}. As the name suggests, \textit{Content Restriction} is to limit the generated words. \textit{Format Restriction} is to standardize the generated  comparative relations format.  \textit{Prompt Words} amis to stimulate the model to generate matching comparative relations.

%===========================
\subsection{Encoder with Prompt Words}
%===========================

To better extract comparative relations, we design a simple yet effective prompt words injection layer. Given an input sentence, $S=[w_1, w_2, \cdots, w_n]$, we use prompt words injection layer to obtain a new input:
\begin{equation}
    S' = [w_1, w_2, \cdots, w_n, t_p],
\end{equation}
where $t_p$ denotes the injected span of prompt words. Then we represent this input by using $\text{Encoder}$:
\begin{equation}
    r = \text{Encoder}([w_1, w_2, \cdots, w_n, t_p]),
\end{equation}
where $r$ refers to the encoded representation of $S'$.

%===========================
\subsection{Comparative Aware Decoder}
%===========================
The comparative aware decoder is built on top of the decoder in GPT-2. The main add-on to the original decoder is a filter layer. Next, we  detail the format of the generated text, then explain the filter layer.

%The remaining is to properly utilize the representation $r$ and the well-designed decoding strategy. To notify the decoder of this information, we design a filter layer except for the original decoder of GPT-2. 

\paratitle{Format of the generated text.}
To better utilize the capability of text generation of pre-trained language models, we formulate our task of comparative relation extraction as a text generation task. 
%Pre-trained language models have significant ability to generate high-quality text, so we transfer the comparative relations to natural language text by a simple rule. 
Specifically, a target comparative relation $(t_1, t_2, a)$ to be extracted from input text, becomes a target piece of text ``$t_1$ vs. $t_2$ in $a$'' to be generated by the model. %This approach allows our model 
Then, we use the decoder to generate text word by word, starting with a [CLS] token (see Figure~\ref{fig:model}):
\begin{equation}
    \mathcal{P}_i = p(r_i|\textbf{r}_{j < i}, S'),
\end{equation}
where $\mathcal{P}_i$ denotes the probability distribution of the predicted $i$-th word, computed based on the input and the earlier generated sequence. The corresponding word $w_i$ is obtained by $\text{argmax}(\mathcal{P}_i)$.

\paratitle{The filter layer.}  A typical generative model generates words from its entire vocabulary. Hence, the model will output words that are not in the original input text. In our problem setting, these words are not part of the expected results. 
% So, when the model generates a relation, it determines whether each element of the relation belongs to the original text. 
To guarantee that all elements of the generated relations are from the original text, we design a filter layer, to filter all relations generated:
\begin{equation}
    S_g = [\cdots, \text{rel}_i, \cdots],
\end{equation}
where $\text{rel}_i$ denotes the $i$-th generated relation and all elements of it (\ie two targets $t_1$ $t_2$, and aspect $a$) must be from the original text $S$. Otherwise, the generated relation is discarded.

%===========================
\subsection{Training Objective}
%===========================

The objective of the \modelname is to predict the transferred comparative relation, from the input text. We adopt objective $L(\theta)$ by simply using the auto-regressive cross-entropy:
\begin{equation}
    L(\theta) = -\sum_{i=1}^{N_c} \log(p(r_i|\textbf{r}_{j < i}, S')),
\end{equation}
where $\theta$ denotes the parameters of the model. $N_c$ represents the total length of the transferred comparative relations.

As discussed earlier, a sentence may contain  multiple comparative relations, making the generation problem complicated. In this paper, we use a simple way to guide the proposed model. For simple comparative relation, the order of targets $t_1$ and $t_2$ follows their relative positions in the original text. If there are multiple relations, we prioritize $t_1$ then $t_2$, if $t_1$ is consistent in multiple relations.

\section{Experiments}
%===========================

\begin{table}
  \centering
  \small{
  \caption{Statistics of the two dataests \camera  and \compsentliu.  The train/dev/test split is $\text{7:1:2}$ for both datasets.}
    \label{tab:datasets}
    \begin{tabular}{r|rl|rl}
    \toprule
    \multicolumn{1}{l|}{Instance} & \multicolumn{2}{l|}{\camera} & \multicolumn{2}{l}{\compsentliu} \\
    \midrule
    \multicolumn{1}{r|}{\multirow{3}[2]{*}{Sentence}} & \multirow{3}[2]{*}{1279} & Train:895 & \multirow{3}[2]{*}{628} & Train:440 \\
          &       & Dev:128 &       & Dev:63 \\
          &       & Test:256 &       & Test:125 \\
    \midrule
    \multicolumn{1}{r|}{\multirow{3}[2]{*}{Relation}} & \multirow{3}[2]{*}{1780} & Train:1218 & \multirow{3}[2]{*}{751} & Train:522 \\
          &       & Dev:172 &       & Dev:74 \\
          &       & Test:390 &       & Test:155 \\
    \bottomrule
    \end{tabular}%
    }

\end{table}%

%===========================
\subsection{Dataset}
\label{sec:exp:dataset}
%===========================

\paratitle{\camera.} This is a manually annotated dataset, containing camera reviews~\cite{DBLP:conf/lrec/KesslerK14}.\footnote{\url{https://wiltrud.hwro.de/research/data/reviewcomp and contanarisons.html}} \camera is currently the largest dataset of comparative sentences, containing $1,279$ comparative sentences and $1,780$ relations. The comparative relation is in the format of (subject, object, opinion words). In this dataset, 74.4\% sentences each contains one relation, 17.4\% sentences contain two relations each, and remaining sentences contain more than two relations each. 

\paratitle{\compsentliu.}  This dataset is  constructed by~\citet{DBLP:conf/coling/GanapathibhotlaL08}, which  contains both comparative and non-comparative sentences.\footnote{\url{https://www.cs.uic.edu/~liub/FBS/sentiment-analysis.html\#datasets}} We select the comparative sentences from this dataset in our experiments. In this dataset, 85.7\% sentences each contains one relation, 11.6\% contain two relations and the remaining contain more than two. 

For both datasets, we randomly split the instances with the ratio of $\text{7:1:2}$ for training, validation/development, and testing. Table~\ref{tab:datasets} reports dataset statistics. 

%The statistics of those two datasets show $80\%$ of the data in the dataset has a comparative relation. Hence, to study the effect of multiple comparative relations, we concatenate the sentences with single relation to imitate multiple relations.
% , we find the number of relations is greater than the number of the sentence. 
% So some sentences have more than one relation. 

%======================
\subsection{Compared Methods}
%======================
We compare \modelname with the following baselines, CRF~\cite{DBLP:journals/comsis/WangXWHL17}, BERT~\cite{DBLP:conf/naacl/DevlinCLT19}, BERT-CRF. All these models are pipeline models: the models extract comparative elements as a sequence labeling task, and then build comparative relations by using Cartesian product. 

We also compare \modelname with GPT-2, the base model of \modelname. Note that, we do not compare our model with traditional rule-based solutions as re-producing such models are time consuming. Due to the small dataset size, we do not train LSTM-based models as well, because LSTM in general requires a lot of training data. 

%as users' writing styles are varied, the prediction of comparative relations~(\ie targets and aspect) in a comparison sentence is a difficult task~\cite{DBLP:conf/cikm/AroraAGP17}. 
%Further, existing comparative datasets in this domain are relatively small. And LSTM generally requires more data to train, so we choose a pre-trained language model instead. BERT~\cite{DBLP:conf/naacl/DevlinCLT19} is used as the baseline because it performs well on few-shot scenarios. We also use CRF~\cite{DBLP:journals/comsis/WangXWHL17}, BERT-CRF~\cite{DBLP:conf/emnlp/LiuXY21} as baseline models. 

%===========================
\subsection{Overall Performance}
%===========================

We use Precision, Recall and $F1$ of extracted relation to evaluate the accuracy of all models. It is worth noting that the two targets in our defined comparative relation are unordered.
% When select three evaluation indexes, including \textit{Relation-pre}, \textit{Relation-rec}, \textit{Relation-$F1$}. 
% In order to unify the evaluation criteria of all models, the output of  all the model is $\{(\{t_1,t_2\},a),(\{t_1,t_2\},a),\dots\}$. When calculating the model outputs, $t_1$ and $t_2$ are unordered. 
An extracted comparative relation is marked as correct, when both targets and the aspect are all correct, otherwise is marked as wrong.
Table~\ref{tab:main_resualt} reports the results of \modelname and 4 baselines, on both datasets.

\begin{table}
  \centering
  \caption{Precision, recall and $F_1$ on both datasets. All  baseline methods are our own implementation.}
%   \caption{Main result on two dataset. GPT-2  refers GPT-2 Fine tuning without prompt words.}
  \small{
    \begin{tabular}{l|ccc}
    \toprule
    \multicolumn{4}{c}{\camera} \\
    \midrule
    Model & Precision & Recall & $F_1$ \\
    \midrule
    CRF   & 0.151  & 0.215  & 0.178  \\
    BERT  & 0.289  & 0.414  & 0.341  \\
    BERT-CRF & 0.279  & \textbf{0.454} & 0.345  \\
    GPT-2  & 0.375 & 0.284  & 0.323  \\
    \modelname  & \textbf{0.419} & 0.329  & \textbf{0.368} \\
    \midrule
    \midrule
    \multicolumn{4}{c}{\compsentliu} \\
    \midrule
    Model & Precision & Recall & $F_1$ \\
    \midrule
    CRF   & 0.000  & 0.000  & 0.000  \\
    BERT  & 0.097  & 0.097  & 0.097  \\
    BERT-CRF & 0.118  & 0.103  & 0.110  \\
    GPT-2  & \textbf{0.275} &	0.213 &	0.240 \\
    \modelname  & 0.267 & \textbf{0.232} & \textbf{0.248} \\
    \bottomrule
    \end{tabular}%
    
    }
  \label{tab:main_resualt}%
\end{table}%

\begin{table}
  \centering
  \small
  \caption{Precision, recall and $F1$ on \camera, by number of comparative relations $N$ in each sentence.}
%  \caption{Split the test dataset of \camera and test the models on different data. N refers to the number of relation in a sentence. R-Pre, R-Rec, R-$F1$ refers to Relation-Pre, Relation-Rec, Relation-F1, respectively. GPT-2 refers GPT-2 model training without prompt words.}
    \begin{tabular}{c|l|ccc}
   \toprule
   
    $N$ & Model &  Precision & Recall & $F1$ \\
    \midrule
     $1$ & CRF   & 0.165  & 0.238  & 0.195  \\
          & BERT  & 0.236  & 0.354  & 0.283  \\
          & BERT-CRF & 0.229  & 0.392  & 0.289  \\
          & GPT-2 & 0.322  & 0.320  & 0.321 \\
          & \modelname& \textbf{0.386} & \textbf{0.392} & \textbf{0.389}\\
      \midrule
    $2$ & CRF  & 0.119  & 	0.138 	& 0.128  \\
          & BERT  & 0.348  & \textbf{0.340} & \textbf{0.344} \\
          & BERT-CRF & 0.198  & 0.277  & 0.231  \\
          & GPT-2 & \textbf{0.422} & 0.287  & 0.342  \\
          & \modelname  & 0.388  & 0.277  & 0.323 \\
    \midrule
    $>2$ & CRF   & 0.152  &	0.245  &	0.187  \\
          & BERT  & 0.341  & 0.588  & 0.432  \\
          & BERT-CRF & 0.428  & \textbf{0.725} & \textbf{0.538} \\
          & GPT-2 & 0.537  & 0.216  & 0.308 \\
          & \modelname & \textbf{0.600} & 0.265  & 0.367   \\
    \bottomrule
    \end{tabular}%
  \label{tab:different_number}%
\end{table}%

%\paratitle{Discussion.} 
On both datasets, our proposed \modelname  achieves the best $F1$. 
On \camera, BERT-based models outperform CRF model,  revealing the powerful ability of pre-trained language models, much as expected. Because BERT-based models first extract comparative elements,  then build relations by combining the extracted elements, these models produce many relations, leading to higher recall but lower precision, compared to generative models GPT-2 and our model.  
On \compsentliu, all pipeline models produce low accuracy, compared to generative models. A key reason is the small dataset size. Sequence labeling methods typically need many instances for training, and \compsentliu does not provide sufficient training data.

%As expected, \modelname performs better than general GPT-2. More interestingly, \modelname outperforms the pipeline model by about $10\%$.

%===========================
\subsection{Number of Comparative Relations}
%===========================

Table~\ref{tab:different_number} reports a detailed look at the results on \camera dataset, a break down of results on sentences containing 1, 2, or more relations. 

\modelname achieves the best scores by all measures, on sentences with a single relation. However, our model does not perform well on sentences containing 2 or more relations. Recall that,  in \camera dataset, majority (or 74.4\%) sentences contain only one relation. Hence the model is learned to focus on one relation extraction only, for a given sentence. 

To more properly evaluate the model capability of handling multiple relations in the input text, we augment the model training by concatenating a pair of sentences  in training set, to form a longer text with multiple relations. Then we evaluate the model on sentences with two or more relations in \camera. The results are reported in Table~\ref{tab:document}. \modelname outperforms all baselines after training with augmented data with multiple relations in input text.

\begin{table}
  \centering
  \small{
  \caption{Precision, recall and $F1$ on texts with two or more comparative relations on \camera dataset.}
%   \caption{Result for multiple sentences on \camera.}
    \label{tab:document}%
    \begin{tabular}{l|ccc}
    \toprule
    Model & Precision & Recall & $F_1$ \\
    \midrule
    BERT  & 0.118  & \textbf{0.455} & 0.188  \\
    BERT-CRF & 0.115  & 0.444  & 0.183  \\
    GPT-2 & 0.335  & 0.237  & 0.277  \\
    \modelname  & \textbf{0.402} & 0.290  & \textbf{0.337} \\
    \bottomrule
    \end{tabular}%
    }

\end{table}%

\begin{table}
  \centering
  \caption{Precision, recall and $F1$ of models with different prompt words on \camera dataset.}
%   \caption{Result of \modelname on  different prompt words.}
  \small{
    \begin{tabular}{l|ccccc}
    \toprule
    Prompt Words & Pre. & Rec. & $F_1$ \\
    \midrule
    ``Let me see:'' & 0.387  & 0.141  & 0.206  \\
    ``[SEP]'' & 0.359  & 0.292  & 0.322  \\
    ``generate relations:'' & 0.367  & 0.289  & 0.323  \\
    ``My name:''& 0.360 &	0.297 &	0.326 \\

    ``relations:'' & 0.393  & 0.316  & 0.350 \\
    ``comparative relations:'' & 0.410  & 0.308  & 0.352 \\
    ``comparative relations tuple:'' & \textbf{0.419} & \textbf{0.329} & \textbf{0.368} \\
    \bottomrule
    \end{tabular}%
    }
  \label{tab:prompt_words}
%   \vspace{-1ex}
\end{table}

%===========================
\subsection{Impact of Prompt Words}
%===========================

Existing studies on prompt-based models reveal that PLM models are heavily affected by the chosen prompt. In our experiments, we also study the impact of different prompt words on our model. Table~\ref{tab:prompt_words} reports the results.

%\paratitle{Discussion.} 
% Different prompt words have different effect on model. 
We list seven popular prompt words in Table~\ref{tab:prompt_words}, including ``generate relation:'', ``Let me see:'', ``My name:'', ``relations:'', ``comparative relation:'', ``[SEP]'' and ``comparative relation tuple:''. 
Among them, prompt ``[SEP]'' cannot guide the model to generate a comparison relationship linguistically. 
% only the role of separating sentences and relations. 
``Let me see:'' and ``My name:'' are less relevant to the task, leading to poor results as expected.
% is similar to ``[SEP]'', but at least it has semantic meaning. 
Prompt words containing ``comparative'' lead to better results than others. The prompt words ``comparative relation tuple:'' achieves the best performance on all measures.

%===========================
\section{Conclusion}
%===========================

In this paper, we study comparative relations extraction and propose \modelname model. The key idea of \modelname is to use generative method to extract complex multiple comparative relations from input text.
Experiment results show that our proposed model achieves state-of-the-art performance against strong baselines. More importantly, \modelname is able to handle texts containing multiple comparative relations, after trained with augmented data.

% Based on the power generation model, we inject prompt words into the sentence. 
% We innovatively use the generation model to directly extract comparative relations. To the best of our knowledge, it is the first approach to jointly extract comparative elements and comparative relations from sentences.

\clearpage
%============================
\section*{Limitations}
%============================

In this paper, we make the very first attempt to perform comparative relation extraction by generative way. 
There are two main limitations: data and model. 
The size of the datasets used in experiments are small, so we cannot design a model with more parameters to perform more in-depth research. Nevertheless, these two are the only two datasets publicly available to our best knowledge. 
Another limitation is the design of the model. This work only considers discrete prompt words and explores a small collection of prompt words to get a local optimum.

For the comparative relation extraction task, an important problem is the lack of comparative elements. Our work and existing studies both ignore this problem. We consider it is an important dimension to explore in future, because sometimes a lacked comparative element may refer to all remaining elements. For example, ``My iPhone 13 is the best''. The missing target here refers to all the remaining phones, and the missing aspect refers to all aspects in comparison.

% From all results above, improving the generation model in multiple relations of a single sentence extraction is critical in the future. What's more, multiple relations of a single sentence account for very few in comparative sentences. Meanwhile, comparative sentence datasets are very small. So, constructing high-quality and sufficient comparative sentences is also an issue to be considered in the future.

\bibliography{anthology,custom}

\begin{thebibliography}{14}
\expandafter\ifx\csname natexlab\endcsname\relax\def\natexlab#1{#1}\fi

\bibitem[{Arora et~al.(2017)Arora, Agrawal, Goyal, and
  Pathak}]{DBLP:conf/cikm/AroraAGP17}
Jatin Arora, Sumit Agrawal, Pawan Goyal, and Sayan Pathak. 2017.
\newblock \href {https://doi.org/10.1145/3132847.3133141} {Extracting entities
  of interest from comparative product reviews}.
\newblock In \emph{Proceedings of the 2017 {ACM} on Conference on Information
  and Knowledge Management, {CIKM} 2017, Singapore, November 06 - 10, 2017},
  pages 1975--1978. {ACM}.

\bibitem[{Devlin et~al.(2019)Devlin, Chang, Lee, and
  Toutanova}]{DBLP:conf/naacl/DevlinCLT19}
Jacob Devlin, Ming{-}Wei Chang, Kenton Lee, and Kristina Toutanova. 2019.
\newblock \href {https://doi.org/10.18653/v1/n19-1423} {{BERT:} pre-training of
  deep bidirectional transformers for language understanding}.
\newblock In \emph{Proceedings of the 2019 Conference of the North American
  Chapter of the Association for Computational Linguistics: Human Language
  Technologies, {NAACL-HLT} 2019, Minneapolis, MN, USA, June 2-7, 2019, Volume
  1 (Long and Short Papers)}, pages 4171--4186. Association for Computational
  Linguistics.

\bibitem[{Duan et~al.(2017)Duan, Tang, Chen, and
  Zhou}]{DBLP:conf/emnlp/DuanTCZ17}
Nan Duan, Duyu Tang, Peng Chen, and Ming Zhou. 2017.
\newblock \href {https://doi.org/10.18653/v1/d17-1090} {Question generation for
  question answering}.
\newblock In \emph{Proceedings of the 2017 Conference on Empirical Methods in
  Natural Language Processing, {EMNLP} 2017, Copenhagen, Denmark, September
  9-11, 2017}, pages 866--874. Association for Computational Linguistics.

\bibitem[{Ganapathibhotla and Liu(2008)}]{DBLP:conf/coling/GanapathibhotlaL08}
Murthy Ganapathibhotla and Bing Liu. 2008.
\newblock \href {https://aclanthology.org/C08-1031/} {Mining opinions in
  comparative sentences}.
\newblock In \emph{{COLING} 2008, 22nd International Conference on
  Computational Linguistics, Proceedings of the Conference, 18-22 August 2008,
  Manchester, {UK}}, pages 241--248.

\bibitem[{Jindal and Liu(2006)}]{DBLP:conf/aaai/JindalL06}
Nitin Jindal and Bing Liu. 2006.
\newblock \href {http://www.aaai.org/Library/AAAI/2006/aaai06-209.php} {Mining
  comparative sentences and relations}.
\newblock In \emph{Proceedings, The Twenty-First National Conference on
  Artificial Intelligence and the Eighteenth Innovative Applications of
  Artificial Intelligence Conference, July 16-20, 2006, Boston, Massachusetts,
  {USA}}, pages 1331--1336. {AAAI} Press.

\bibitem[{Kessler and Kuhn(2013)}]{DBLP:conf/emnlp/KesslerK13}
Wiltrud Kessler and Jonas Kuhn. 2013.
\newblock \href {https://aclanthology.org/D13-1194/} {Detection of product
  comparisons - how far does an out-of-the-box semantic role labeling system
  take you?}
\newblock In \emph{Proceedings of the 2013 Conference on Empirical Methods in
  Natural Language Processing, {EMNLP} 2013, 18-21 October 2013, Grand Hyatt
  Seattle, Seattle, Washington, USA, {A} meeting of SIGDAT, a Special Interest
  Group of the {ACL}}, pages 1892--1897. {ACL}.

\bibitem[{Kessler and Kuhn(2014)}]{DBLP:conf/lrec/KesslerK14}
Wiltrud Kessler and Jonas Kuhn. 2014.
\newblock \href
  {http://www.lrec-conf.org/proceedings/lrec2014/summaries/1001.html} {A corpus
  of comparisons in product reviews}.
\newblock In \emph{Proceedings of the Ninth International Conference on
  Language Resources and Evaluation, {LREC} 2014, Reykjavik, Iceland, May
  26-31, 2014}, pages 2242--2248. European Language Resources Association
  {(ELRA)}.

\bibitem[{Liu et~al.(2021)Liu, Xia, and Yu}]{DBLP:conf/emnlp/LiuXY21}
Ziheng Liu, Rui Xia, and Jianfei Yu. 2021.
\newblock \href {https://doi.org/10.18653/v1/2021.emnlp-main.322} {Comparative
  opinion quintuple extraction from product reviews}.
\newblock In \emph{Proceedings of the 2021 Conference on Empirical Methods in
  Natural Language Processing, {EMNLP} 2021, Virtual Event / Punta Cana,
  Dominican Republic, 7-11 November, 2021}, pages 3955--3965. Association for
  Computational Linguistics.

\bibitem[{Radford et~al.(2019)Radford, Wu, Child, Luan, Amodei, Sutskever
  et~al.}]{radford2019language}
Alec Radford, Jeffrey Wu, Rewon Child, David Luan, Dario Amodei, Ilya
  Sutskever, et~al. 2019.
\newblock Language models are unsupervised multitask learners.
\newblock \emph{OpenAI blog}, 1(8):9.

\bibitem[{Tu et~al.(2022)Tu, Li, Cui, Wang, Wen, and
  Yan}]{DBLP:conf/acl/TuLC0W022}
Quan Tu, Yanran Li, Jianwei Cui, Bin Wang, Ji{-}Rong Wen, and Rui Yan. 2022.
\newblock \href {https://aclanthology.org/2022.acl-long.25} {{MISC:} {A} mixed
  strategy-aware model integrating {COMET} for emotional support conversation}.
\newblock In \emph{Proceedings of the 60th Annual Meeting of the Association
  for Computational Linguistics (Volume 1: Long Papers), {ACL} 2022, Dublin,
  Ireland, May 22-27, 2022}, pages 308--319. Association for Computational
  Linguistics.

\bibitem[{Tu et~al.(2016)Tu, Lu, Liu, Liu, and Li}]{DBLP:conf/acl/TuLLLL16}
Zhaopeng Tu, Zhengdong Lu, Yang Liu, Xiaohua Liu, and Hang Li. 2016.
\newblock \href {https://doi.org/10.18653/v1/p16-1008} {Modeling coverage for
  neural machine translation}.
\newblock In \emph{Proceedings of the 54th Annual Meeting of the Association
  for Computational Linguistics, {ACL} 2016, August 7-12, 2016, Berlin,
  Germany, Volume 1: Long Papers}. The Association for Computer Linguistics.

\bibitem[{Wang et~al.(2017)Wang, Xin, Wang, Huang, and
  Liu}]{DBLP:journals/comsis/WangXWHL17}
Wei Wang, Guodong Xin, Bailing Wang, Junheng Huang, and Yang Liu. 2017.
\newblock \href {https://doi.org/10.2298/CSIS161229031W} {Sentiment information
  extraction of comparative sentences based on {CRF} model}.
\newblock \emph{Comput. Sci. Inf. Syst.}, 14(3):823--837.

\bibitem[{Zhang et~al.(2021)Zhang, Luan, Sun, Zhai, Xu, and
  Liu}]{DBLP:journals/taslp/ZhangLSZX021}
Jiacheng Zhang, Huanbo Luan, Maosong Sun, Feifei Zhai, Jingfang Xu, and Yang
  Liu. 2021.
\newblock \href {https://doi.org/10.1109/TASLP.2021.3057831} {Neural machine
  translation with explicit phrase alignment}.
\newblock \emph{{IEEE} {ACM} Trans. Audio Speech Lang. Process.},
  29:1001--1010.

\bibitem[{Zhang et~al.(2020)Zhang, Sun, Galley, Chen, Brockett, Gao, Gao, Liu,
  and Dolan}]{DBLP:conf/acl/ZhangSGCBGGLD20}
Yizhe Zhang, Siqi Sun, Michel Galley, Yen{-}Chun Chen, Chris Brockett, Xiang
  Gao, Jianfeng Gao, Jingjing Liu, and Bill Dolan. 2020.
\newblock \href {https://doi.org/10.18653/v1/2020.acl-demos.30} {{DIALOGPT} :
  Large-scale generative pre-training for conversational response generation}.
\newblock In \emph{Proceedings of the 58th Annual Meeting of the Association
  for Computational Linguistics: System Demonstrations, {ACL} 2020, Online,
  July 5-10, 2020}, pages 270--278. Association for Computational Linguistics.

\end{thebibliography}
\bibliographystyle{acl_natbib}
\end{document}